\documentclass[11pt,a4paper]{article}
\usepackage[hyperref]{emnlp-ijcnlp-2019}
\usepackage{times}
\usepackage{latexsym}

\usepackage{url}

\usepackage{graphicx}
\usepackage{amsmath}
 \usepackage{subfigure}
 \usepackage[export]{adjustbox}
  
\aclfinalcopy % Uncomment this line for the final submission

\title{Concept Pointer Network for Abstractive Summarization}
\author{ Wang Wenbo$^1$, Gao Yang$^{1,2}$\thanks{\,\,  Corresponding author} $\,\,$, Huang Heyan$^{1,2}$ and Zhou Yuxiang$^1$\\
{ $^1$ School of Computer Science and Technology}\\{ Beijing Institute of Technology, Beijing, China, 100081}\\
{ $^2$ Beijing Engineering Research Center of High Volume Language}\\{
Information Processing and Cloud Computing Applications}\\
{\tt gyang@bit.edu.cn}\\
}

\date{}
\begin{document}
\maketitle
\begin{abstract}
  A quality abstractive summary should not only copy salient source texts as summaries but should also tend to generate new conceptual words to express concrete details. Inspired by the popular pointer generator sequence-to-sequence model, this paper presents a concept pointer network for improving these aspects of abstractive summarization. The network leverages knowledge-based, context-aware conceptualizations to derive an extended set of candidate concepts. The model then points to the most appropriate choice using both the concept set and original source text. This joint approach generates abstractive summaries with higher-level semantic concepts. The training model is also optimized in a way that adapts to different data, which is based on a novel method of distantly-supervised learning guided by reference summaries and testing set. Overall, the proposed approach provides statistically significant improvements over several state-of-the-art models on both the DUC-2004 and Gigaword datasets.  A human evaluation of the model's abstractive abilities also supports the quality of the summaries produced within this framework. 
\end{abstract}

 \section{Introduction}

Abstractive summarization (ABS) has gained overwhelming success owing to a tremendous development of sequence-to-sequence (seq2seq) model and its variants \cite{DBLP:conf/emnlp/RushCW15,Chopra2016Abstractive,paulus2017deep,DBLP:conf/acl/BansalPG18,Gao2019}. 
%. Especially, the seq2seq and its variants have been dominantly  accepted by the community of machine translation, dialog generation, and also been successfully applied in abstractive summarization (ABS) . 
In tandem with seq2seq models, pointer generator was developed by  \citet{DBLP:conf/acl/SeeLM17} as a solution to tackle the rare words and out-of-vocabulary (OOV) problem associated with generative-based models. The idea behind is to use attention as a pointer to determine the probability of generating a word from both a vocabulary distribution and the source text. 
%To tackle the rare words and out-of-vocabulary (OOV) problem in the generative-based models, pointer network \cite{vinyals2015pointer}  was developed to use attention as a pointer to decide the probability of  generating a word from both a vocabulary distribution and a  source text \cite{DBLP:conf/acl/SeeLM17}. 
%However, rare words and out-of-vocabulary (OOV) has always been a problematic issue for these  generative-based models.  Copy mechanism was proposed to copy certain information from source texts to eliminate the OOV problem. 
%Further, pointer generator network \cite{DBLP:conf/acl/SeeLM17} generated summaries from both of source texts and overall vocabulary distribution, which took an advantage of probabilistic  balance between them. 
Pointer generator networks have also been extensively accepted by the ABS community due to their efficacy with long document summaries \cite{chen2018fast,hsu2018unified}, title summarization \cite{sun2018multi}, etc.
%first the drawbacks of copynet, the development of pointer-generator, how it works. 
 
\begin{figure}[t]
	\centering
	\includegraphics[width=0.5\textwidth]{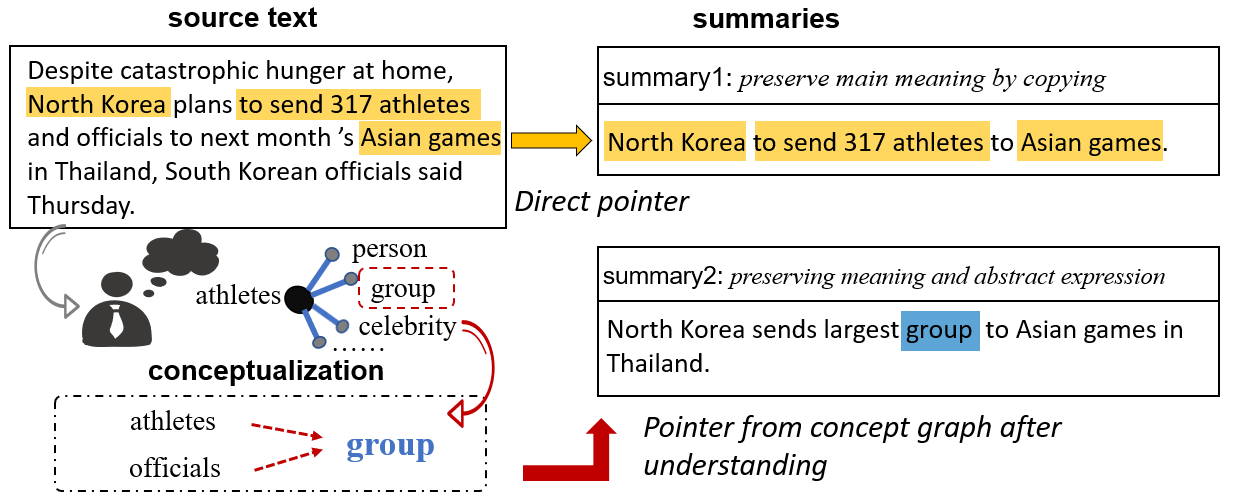}
	\caption{``summary1" only copies keyword from the source text, while ``summary2"  generates new concepts to convey the meaning.}  
	\label{fig:intro}
\end{figure}
%However, the current power of abstractive summarization falls short of their potential. A good summary should not simply copy original material, it should also generate new and even abstract concepts that reflect high-level semantics. As the example in Figure \ref{fig:intro} shows, a more human-like summary would be based on one's own understanding of the detail in the words, expressed as higher-level concepts drawn from world knowledge - like using the word ``group " to replace ``athletes and officials". Conversely, a seq2seq model with a pointer mechanism (marked as the direct pointer in the figure) is likely to merely copy parts of the original text to form a summary using keywords and phrases, such as ``317 athletes".

However, the current power of abstractive summarization falls short of their potential. As the example in Figure \ref{fig:intro} shows, a seq2seq model with a pointer mechanism (marked as the direct pointer) is likely to merely copy parts of the original text to form a summary using keywords and phrases, such as ``317 athletes". 
Conversely,  a more human-like summary would be based on one's own understanding of the detail in the words, expressed as higher-level concepts drawn from world knowledge\textemdash like using the word ``group" to replace ``athletes and officials".
 This indicates that a good summary should not simply copy original material, it should also generate new and even abstract concepts that reflect high-level semantics.

%has a typical  characteristics that is not just to copy what it originally exists, but also produce new and even abstract concepts with higher level of semantics.  
%As an example shown in Figure \ref{fig:intro}, human make a summary  based on his/her own understanding of those detailed words and then summarize them with high-level concepts. Concretely, a seq2seq model with pointer mechanism, which is marked as direct pointer in the figure, is likely to copy original texts to the summary, involving detailed words  such as ``317 athletes"; while human is able to understand the detailed content and then point to a correct concept from world knowledge as a summary, such as using ``group" to summarize the ``athletes and  officials". 

Therefore, a pointer generator network that solely considers the source material to generate a summary does not adequately satisfy the needs of high-quality  abstractive summarization. We argue that  concepts have a greater ability to  express deeper meanings  than verbatim words. 
As such, it is essential to explore the potential of using concepts from world knowledge to assist with abstractive summarization. 
Our developed model not only points to informative source texts but also  leverages conceptual words from human knowledge in the summaries it generates.  

%in addition to pointing salient information from original texts, 

Hence, in this paper, we propose a novel model based on a concept  pointer generator that  encourages the generation of conceptual and abstract words. As a hidden benefit, the model also alleviates the OOV problems. Our model uses pointer network to capture the salient information from a source text, and then employs another pointer to generalize the detailed words according to their upper level of expressions. Finally, the output is also consistent with language model by the seq2seq generator. 
Unique to our concept pointer is a set of concept candidates particular for a word that is drawn from a huge knowledge base. The set of candidates adheres to a concept distribution, where the probability of each concept being generated is linked to how strongly the candidate represents each word. 
Moreover, the concept distribution is iteratively  updated to better explain the target word given the context of the source material and inherent semantics in the texts. Hence, the learned concept pointer points to the most suitable and expressive concepts or words. The optimization function is adaptive so as to cater for different datasets with distantly-supervised training. The network is then optimized end-to-end using reinforcement learning, with the distant-supervision strategy as a complement to further improve the summary.

%and adhere with a concept distribution for the word where each probability represents how strongly the word belongs to each concept. On account of the word's contexts and the  semantics itself, the concept distribution is accordingly updated for better explaining the word. As such, the learned concept pointer takes effect on pointing to those suitable and expressive concepts. Moreover, we propose to optimize the model for better adaptive to different datasets with a distantly supervised training.  
%The network is then optimized end-to-end using reinforcement learning and  the proposed distant supervision strategy to achieve reasonable improvements, respectively.

Overall, the contributions of this paper are: 1) a novel concept pointer generator network that leverages context-aware conceptualization and a concept pointer, both of which are jointly integrated into the generator to deliver informative and abstract-oriented summaries; 2) a novel distant supervision training strategy that favors model adaptation and generalization, which results in performance that outperforms the well-accepted evaluation-based reinforcement learning optimization on a test-only dataset in terms of ROUGE metrics; 3) a statistical analysis of quantitative results and human evaluations from comparative experiments with several state-of-the-art models that shows the proposed method provides promising performance.

\section{Related Work}
\label{sec:relatedwork}
%Text summarization techniques can be divided into extractive and abstractive approaches. Extractive summarization has the advantage of producing fluent and readable summaries because it directly selects key segments of the given material. However, extractive approaches do not produce noise-free summaries, and the models do not fully understand the source content. 
 
Abstractive summarization supposedly digests and understands the source content and, consequently, the generated summaries are typically a reorganization of the wording that sometimes form new sentences. 
%It is complex and represents the ultimate goal of any summarization system. 
Historically, abstractive summarization has been performed through rule-based sentence selection \cite{dorr2003hedge}, key information extraction  \cite{DBLP:conf/acl/GenestL11}, syntactic parsing \cite{DBLP:conf/acl/BingLLLGP15} and so on. However, more recently, seq2seq models with attention have played a more dominant role in generating abstractive summaries \cite{DBLP:conf/emnlp/RushCW15,Chopra2016Abstractive,DBLP:conf/conll/NallapatiZSGX16,DBLP:conf/acl/ZhouYWZ17}.  Extensions to the seq2seq approach include an intra-decoder attention \cite{paulus2017deep} and coverage vectors  \cite{DBLP:conf/acl/SeeLM17} to decrease repetition in phrasing. 
%CGU sets a convolutional gated unit for global encoding to select relevant phrases  \cite{DBLP:conf/acl/LinSMS18}. 
Copy mechanism \cite{gu2016incorporating}
%and pointer networks \cite{DBLP:conf/acl/SeeLM17} 
has been integrated into these models to tackle OOV problem.  
\citet{DBLP:conf/aaai/ZhouYWZ18} went on to propose SeqCopyNet  which copies  complete sequences from an input sentence to further maintain the readability of the generated summary.

Pointer mechanism \cite{vinyals2015pointer} has drawn 
much attention in text summarization \cite{DBLP:conf/acl/SeeLM17}, because this technique not only provides a potential solution for rare words and OOV  but also  extends abstractive summarization in a flexible way \cite{DBLP:conf/naacl/CelikyilmazBHC18}. 
%It can decide whether to generate a token from a vocabulary distribution or from the source text. Plus, this mechanism  can be combined with multiple agents  \cite{DBLP:conf/naacl/CelikyilmazBHC18} to collaboratively generate final vocabulary  distribution.
Further, pointer generator models can effectively adaptive to both extractor and abstractor networks \cite{chen2018fast},  
 %inconsistency loss \cite{hsu2018unified}\\
and summaries can be generated by incorporating a pointer-generator and multiple relevant tasks \cite{DBLP:conf/acl/BansalPG18}, such as question or entailment generation, or  multiple source texts \cite{sun2018multi}.

%Different training strategies have also been used to  improve the ROUGE score (Paulus et al., 2017) or textual
However, work particularly targets the problem of the abstraction is rare. Abstract Meaning Representation (AMR) is used to transform a sentence into a concept graph, then merge those similar concept nodes to form a new summary graph \cite{liu2018toward}. Concepts are also incorporated as auxiliary features \cite{guo2017conceptual}.  \citet{DBLP:conf/emnlp/KryscinskiPXS18} and \citet{weber2018controlling} define the number of new $n$-grams  as the primary criteria of abstractiveness. This makes sense in most cases. But, we believe that abstraction means summarizing detailed content with higher-level semantically related concepts, which has motivated the development of the model proposed in this paper.
%But, we believe that the abstraction means summarizing detailed content with a semantic-related concept in our proposed model.  % 定义 

\section{The Proposed Model}

% Input SET X, n dimension, i step,
% Output SET Y, m dimension, t step,
% concept SET C, k NO., j-th,  
% Context vector h*_i, 
% pointer generator probability, which is probability determines whether to generate a word ,  p_g
% the probability from the vocabulary P_{voc}
% RL, r
\begin{figure*}[t]
	%\centering
	\includegraphics[width=0.9\textwidth,right]{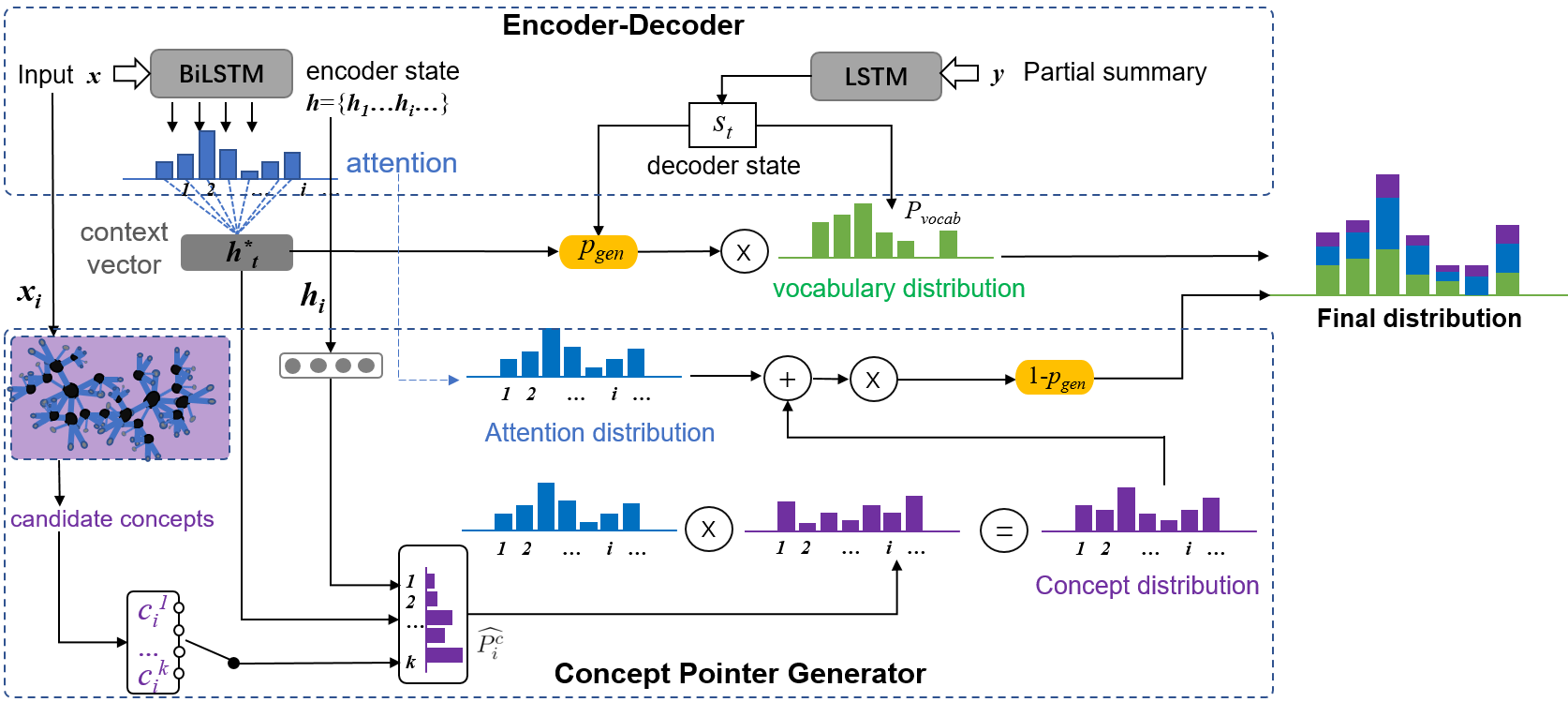}
	%[width=0.65\textwidth]{framework.png}
	\caption{The architecture of our model. Blue bar represents the attention distribution over the inputs. Purple bar represents the concept distribution over the inputs. Noted that, this distribution can be sparse since not every word has its upper concept. Green bar represents the vocabulary distribution generated from seq2seq component.}  
	\label{fig:framework}
\end{figure*}

Neural abstractive summarization can be described as a  generation process where a sequential input is summarized into a shorter sequential output through a neural network.  Suppose that the sequential  input  $\boldsymbol{x} = \lbrace  {x}_1, \dots,  {x}_i, \dots,   {x}_n \rbrace$ is a sequence of  $n$ number of words, and $i$ is the index of the input.  The shorter (i.e., summarized)  sequence of output  is denoted as  $\boldsymbol{y} = \lbrace  {y}_1, \dots,  {y}_t, \dots,   {y}_m \rbrace$ with number of $m$ words, and $t$ indicates a time step.  
%$\boldsymbol{x}$ is fed into the model with the aim to understand the input and grasp main idea of it, and then  produce the output $\boldsymbol{y}$ sequentially. 
As Figure \ref{fig:framework} shows, our model consists of two sub-modules \textemdash an encoder-decoder module and the proposed concept pointer generator module.  

\subsection{Encoder-Decoder Framework}
This process is formulated as an encoder-decoder framework that  
consists of an encoder and an attention-equipped decoder. 
%The encoder maps the original text to a vector and the decoder transforms the vector to a summary.  
We use a two-layer bi-directional LSTM-RNN encoder and  one-layer uni-directional LSTM-RNN decoder  along  with attention mechanism.  

Formally, the encoder  produces sequential hidden states as $(\overrightarrow{\boldsymbol{h}}_1, \dots, \overrightarrow{\boldsymbol{h}}_n)$  and $(\overleftarrow{\boldsymbol{h}}_1, \dots, \overleftarrow{\boldsymbol{h}}_n)$  in the corresponding positions, and the  bi-directional ${h}_i = f_{LSTM}({h}_{i-1},{x}_i)$. Each word $x_i$ in the sequence can be   represented as a concatenation of  the bi-directional hidden states, i.e., $\boldsymbol{h}_i = \lbrack \overrightarrow{\boldsymbol{h}}_i, \overleftarrow{\boldsymbol{h}}_i \rbrack$. 
%As we can see from the structure, the hidden representation of each word contains the context information  from both directions.
The decoder  generates a  target summary from a vocabulary distribution $P_{vocab}(w)$, which is based on a context vector $\boldsymbol{h}_t^*$ through the following  process:
\begin{equation}
\label{eq:outy}
\begin{split}
P_{vocab}(w) & =P(y_t|\boldsymbol{y}_{<t},\boldsymbol{x}; \theta ) \\
 =& \text{sfm}(\boldsymbol{W}_2(\boldsymbol{W}_1[\boldsymbol{s}_t, \boldsymbol{h}^*_t]+\boldsymbol{b}_1)+\boldsymbol{b}_2)
\end{split}
\end{equation}
where  $\boldsymbol{s}_t$ is the hidden state of the decoder at time step $t$ , and $\boldsymbol{h}^*_t$ is the context vector at time step $t$. $\boldsymbol{W}_1, \boldsymbol{W}_2$, $\boldsymbol{b}_1, \boldsymbol{b}_2$ are trainable parameters, and  sfm($\cdot$) is short for softmax function. 
 
The context vector $\boldsymbol{h}^*_t$ is computed by a weighted sum of the hidden  representations of the source text, and the weight is denoted as the attention ${\alpha}_{t,i}$. %  when generating the output summary  
\begin{equation}
\begin{split}
\boldsymbol{h}^*_t &= \sum\nolimits_{i=1}^{n}{\alpha}_{t,i}\boldsymbol{h}_i\\
\alpha_{t,i} &= \text{sfm}(\boldsymbol{v}^\top\tanh(\boldsymbol{W}_h \boldsymbol{h}_i + \boldsymbol{W}_s \boldsymbol{s}_t + \boldsymbol{b}))
\end{split}
\label{equ:contex}
\end{equation}
The softmax function  normalizes the vector of a distribution over
the input position, and $\boldsymbol{v}, \boldsymbol{W}_h, \boldsymbol{W}_s, \boldsymbol{b}$ are trainable parameters.

\subsection{Concept Pointer Generator}
 Pointer networks use attention as a pointer to select segments of the input as outputs \cite{vinyals2015pointer}. As such, a pointer network is a suitable mechanism for extracting salient information, while remaining flexible enough to interface with a seq2seq model for generating an abstractive summarization  \cite{DBLP:conf/acl/SeeLM17}. Our proposed model is essentially an upgrade to this configuration that integrates a new concept pointer network within a unified framework.

 %linking a word to its context-appropriate concept refers to conceptualization in the field of knowledge representation.

\subsubsection{Context-aware Conceptualization}
``Understanding" the instances of a word requires a taxonomic knowledge base that relates those words to a concept space.  In our model, we use an $isA$ taxonomy, called the Microsoft Concept Graph\footnote{The Microsoft concept graph was derived from Probase project. The public data can be downloaded via the provided API:  \url{https://concept.research.microsoft.com/Home/API}} \cite{wang2015inference}, to serve this purpose for two reasons \cite{understanding-short-texts}. First, this graph provides a huge concept space with multi-word terms that cover concepts of worldly facts as concepts, instances, relationships, and values\footnote{The current version is mined from billions of web pages, containing 5.3 M unique concepts, 12.5  M unique entities and 85 M $isA$ relations.}.  Second, the relationships between concepts and entities are probabilistic as a measure of how strongly they are related. Moreover, the probabilities are trustworthy given they have been derived from evidence found in billions of webpages, search log data, and other existing taxonomies. Our model is data-driven and, therefore, is more easily adaptable with probabilities. All these characteristics make the Microsoft Concept Graph a suitable choice for our model.  %are in line with the neural-based abstractive summarization.
More detailed examples are available in Appendix A.

The concept graph specifies the probability that each instance $x$ belongs to a  concept $c$, $p(c|x)$. Given a word $x$, we have a distribution over a set of related concepts. Yet, this raises the question of how to identify a context-appropriate concept for a word from the distributional set of candidate concepts. For instance, {\em apple} in the context of ``{\em an apple is good for you health}" tends to be associated with the concept of {\em fruit} instead of {\em company}. Formally, given a  word $x_i$ in a training sentence, a set of $k$ concept candidates, $C_i = \{c_i^1, c_i^2,\cdots, c_i^k\}$, is linked to the word from the knowledge base, with distributional  probabilities over the concepts, i.e., $P(C|x_i) =\{ p(c_i^1), p(c_i^2), \cdots, p(c_i^k) \}$. The task is to find the most suitable concept $c_i^j$ to fit the updated context, represented by the  vector $\boldsymbol{h}_t^*$ in Equation (\ref{equ:contex}), at each time step $t$. 

In the case of generating summaries given updated contexts,  a weighted update of the distributional concept candidates needs to be  performed. In the model, the updated weight, denoted as $\beta_i^j$, is estimated by a softmax classifier that is jointly conditioned on the hidden representation of the word $\boldsymbol{h}_i$, the context vectors $\boldsymbol{h}^*_t$, and each of concept  vectors:
% \begin{equation}
%      \beta_i^j = \text{sfm}(\boldsymbol{W}_{h'} \boldsymbol{h}_i  + \boldsymbol{W}_{h_t} \boldsymbol{h}^*_t + \boldsymbol{W}_{c}\boldsymbol{c}_i^j)
% \end{equation}
\begin{equation}
     \beta_i^j = \text{softmax}(\boldsymbol{W}_{h'}[ \boldsymbol{h}_i, \boldsymbol{h}^*_t,  \boldsymbol{c}_i^j])
\end{equation}
where $j\in[1,k]$, $\boldsymbol{W}_{h'}$  %\boldsymbol{W}_{h_t}, \boldsymbol{W}_{c}$ 
is a trainable parameter, and $\boldsymbol{c}_i^j$ is the vector of the  $j$th concept candidate, which is a representation of the input embeddings. 

Together with the concept probability from the existing knowledge base $p(c_i^j)$ and the updated weights based on the contexts $\beta_i^j$, a context-aware conceptualized probability of $j$th concept for the word $x_i$,  $P_{i,j}^c$,  is finally estimated as 
\begin{equation}
\label{p_ijc}
    P_{i,j}^c = p(c_i^j)+ \gamma \beta_i^j 
\end{equation}
where $\gamma$ is a tunable parameter.  
Theoretically,  
we will end up with a number of $k$ relevant concepts for each word  $C_i = \{c_i^1, \cdots, c_i^k\}$ with a probability distribution over the set, which is learned as $P^c_i = \{ P_{i,1}^c, \cdots ,  P_{i,j}^c, \cdots, P_{i,k}^c\}$.
%wwen 上述描述有误，应该是这样的：(1)从“\beta_i^j”(j的范围是[1,k])中找到最大值(假设是\beta_i^a最大，其中a属于[1,k])。(2)根据a值找到概念词c_i^a(c_i^a为单词x_i最终的概念词)。(3)根据概念词c_i^a找到其在微软概念图中的输出概率p(c_i^a)。(4)我们认为p(c_i^a)并不能够完全代表c_i^a在文本中作为单词x_a的概念词的概率，需要用\beta_i^a进行修正，公式为P_{i,a}^c = p(c_i^a)+ \gamma \beta_i^a。然后将P_{i,a}^c加入到公式(8)中，代替\widehat{P_{i}^c}。//进一步解释：上述注释内容为从候选词中选择概率词的过程，按照概率最大原则。因此原论文中公式（6）需要修改。
%wwen 对于随机选择概念词的方法，原论文中描述正确。如果需要和按照最大概率原则方法流程尽可能相同，则流程如下：从“\beta_i^j”(j的范围是[1,k])中随机选择一个概率(假设选择了\beta_i^s最大，其中s属于[1,k])。后续步骤和上述注释内容相同。

\subsubsection{Concept Pointer Generator}
The basic pointer generator network contains two sub-modules, one is the pointer network and the other is the generation network. These two sub-modules jointly determine the probabilities of the words in the final generated summary. The generation probability $p_{gen}$  for the generation network \cite{DBLP:conf/acl/SeeLM17} is learned  by 
\begin{equation}
    p_{gen} = \sigma(\boldsymbol{W}_{h^*}\boldsymbol{h}^*_t+\boldsymbol{W}_s \boldsymbol{s}_t +\boldsymbol{W}_y\boldsymbol{y}_{t-1}+\boldsymbol{b}_{gen})
\end{equation}
where $\sigma$ is a sigmoid function.  

For the pointer network, our model consists of a pointer to the source text and a further concept pointer to the relevant concepts that have arisen from the source content. These two separate pointers are calculated as follows. The first pointer is taken based on the attention distribution $\alpha_{t,i}$ over the source text. The second concept pointer is operated over a concept distribution of the source text that is scaled element-wise by the attention distribution. 

To train the model, 
%one specific concept is the need for the update process. 
given the likelihood of each concept in the current context, the updates could be performed in two ways. In a hard assignment, the concept that receives the highest score would be selected for the update:
% \begin{equation}
% \label{equ:max}
% \widehat{P_{i}}^c_\text{argmax} = \arg\max_{j} (P_{i,j}^c|P_{i,j}^c\in P_i^C)
% \end{equation}
 \begin{equation}
% \begin{split}
\label{equ:max}
 \widehat{P_{i}^c}_\text{argmax} = P_{i,a}^c, \text{where } a = \arg\max_{j} (\beta_i^j) 
% \end{split}
\end{equation}
where $a$ is the index of maximized generated weight based on the  contexts, and $P_{i,a}^c$ is obtained by Eqs. (\ref{p_ijc}).

In random selection, each of the concept candidates could be trained randomly to update the parameters: 

\begin{equation}
\label{equ:random}
    \widehat{P_{i}^c}_\text{random} = P_{i,j}^c \sim P_i^C
\end{equation}
where $j$ represents the selected concept index. Considering the above baseline generation network and both the  pointer networks, our final  output distribution is 
\begin{equation}
\begin{split}
    &P_{\text{final}}(w) = p_{gen}P_{vocab}(w)+ (1-p_{gen})\\ 
    &(\sum\nolimits_{i:w_i=w}\alpha_{t,i}+ \sum\nolimits_{i:w_i=w} \alpha_{t,i}\times\widehat{P_{i}^c}) 
\end{split}
\end{equation}
where $\widehat{P_i^c}$ can be updated by $\widehat{P_{i}^c}_{\arg\max}$, or $\widehat{P_{i}^c}_\text{random}$. The difference between these two choices is demonstrated  in the  Experiments section. 

\subsection{Objective Learning}
\subsubsection{Basic MLE}
The baseline objective is derived by maximizing the likelihood training for the seq2seq generation, given a reference summary $y^* = \{y^*_1, y^*_2,\cdots, y^*_{m'}\}$ for document $x$. The training objective is to 
minimize the negative log-likelihood of the target word
sequence:
\begin{equation}
\resizebox{\hsize}{!}{
   $ \mathcal{L}_{MLE} = -\sum_{t = 1}^ {m'} \log P(y_t^*| y_1^*,\cdots, y_{t-1}^*, \boldsymbol{x}) $
    \label{equ:mle}
    }
\end{equation}

\subsubsection{Evaluation based Reinforcement Learning (RL)}
Similar to \citet{paulus2017deep}, policy gradient methods can directly optimize discrete target evaluation metrics, such as ROUGE. 
%At each time step, the word generation is treated as an action taken by the RL agent.
The basic idea is to explore new sequences and compare them to the
best greedy baseline sequence. Once the baseline sequence $\hat{y}$,  or sampled sequence $y^s$, are  generated, they are compared against the reference sequence $y^*$ to compute the rewards $r(\hat{y})$ and $r(y^s)$, respectively. In the RL training stage, two separate output candidates at each time step are produced:  $y^s$ is sampled from the probability distribution $P(y^s_t|y^s_1,\cdots,y^s_{t-1},x)$ %at each time step
, and $\hat{y}$ is the baseline output. 
%, which is greedily generated by maximizing decoding from the vocabulary probability distribution at each time step. 
The training objective is then to minimize
\begin{equation}
\resizebox{\hsize}{!}{
    $\mathcal{L}_{RL} = (r(\hat{y})-r(y^s)) \sum_{t = 1}^{m'} \log P(y_t^s|y^s_1,\cdots,y^s_{t-1},x)$
    }
\end{equation}

It is noteworthy that the samples $y^s$ are selected from a wide range of vocabularies extended by all the concept candidates. This strategy ensures that the model learns to generate sequences with  higher rewards by better exploring a set of close concepts.  

Thus, the combination of these two objectives yield improved
task-specific scores while catering a better language model: 
$
    \mathcal{L}_{final} = \lambda \mathcal{L}_{RL} + (1-\lambda) \mathcal{L}_{MLE}
$, 
where $\lambda$ is a soft-switch between the two objectives.  The model  is pre-trained with MLE loss, then switch to the final loss. 

\subsubsection{Distant Supervision (DS) for Model Adaption} 

%Our intuition is that, if the semantics-dissimilar labeled summary-document pairs adaptively weaken the influence on final loss, the better the fit of the  training model with specific testing data. However, given there are no explicit supervision labels to indicate whether the training set is close to testing set, a new training regime is needed. In answer to this need and also to provide end-to-end functionality in the model, we developed a simple approach for labeling summary-document pairs by calculating the Kullback-Leibler (KL) divergence between each training reference summary and a set of testing documents. In this way, the training pairs are distantly-labelled. %  for training the model.  

Our intuition is that, if the summary-document pairs are dissimilar to the testing set, the model  could be retrained to adapt  to weaken the influence of the dissimilarity on the final loss.   
The result would be a training model that better fits the specific testing data. The challenge is that there are no explicit supervision labels to indicate whether the training set is close to the testing set, so a new training paradigm is needed. In answer to this need and also to provide end-to-end functionality in the model, we developed a simple approach for labeling summary-document pairs by calculating the Kullback-Leibler (KL) divergence between each training reference summary and a set of testing documents. In this way, the training pairs are distantly-labelled for training the model.

Specifically, the representations of the reference summaries and the testing set are computed by summing all the involved word embeddings. Given a testing document $x^{\text{d}}_l$, where $l\in [1, N^{\text{d}}]$ and $N^{\text{d}}$ is the size of the testing corpus, the vector-based representation of one document is  $\boldsymbol{x}_l^{\text{d}} = \exp (\sum_{i=1}^{n'}\boldsymbol{x}^{\text{d}}_i$), where $n'$ is the number of document words involved. The reference summary is represented by $\boldsymbol{y^*} =\exp (\sum_{t=1}^{m'}\boldsymbol{y^*_t}$). We  normalize these vectors through a softmax function to cater for KL calculation. The model adaption with the distant labels is defined as:
%wwen:在上面段落中:第1点:$\boldsymbol{x}_l^{test} = \sum_{i=1}^{n'}\boldsymbol{x}^{test}_i$应该改为$\boldsymbol{x}_l^{test} = exp(\sum_{i=1}^{n'}\boldsymbol{x}^{test}_i)$。第2点:$\boldsymbol{y^*} =\sum_{t=1}^{m'}\boldsymbol{y^*_t}$应该改为$\boldsymbol{y^*} =exp(\sum_{t=1}^{m'}\boldsymbol{y^*_t})$。也就是在上述两个公式前面加入指数函数。

\begin{equation}
\resizebox{\hsize}{!}{
   $ \mathcal{L}_{DS} = (\pi -\frac{1}{N^{\text{d}}}\sum_{l=1}^{N^{\text{d}}} D_{\text{KL}}(\boldsymbol{y^*},\boldsymbol{x}_l^{\text{d}})) \mathcal{L}_{\text{MLE}}$
    }
\end{equation} 
where $D_\text{KL}(.)$ indicates the KL divergence  between $y^*$ and $x_l^{\text{d}}$, and $\pi$ is a constant parameter that is tuned via adaption to the testing set. The divergences are averaged within the testing set, which indicates the overall distances between testing set and each of the reference summary-document pairs.
In this way, the samples in the training corpus are distantly annotated as either relevant or irrelevant for model adaption, noting that the model is pre-trained with the MLE loss before switching to distantly-supervised training. 
%We adopt ROUGE$_\text{F1}$, a text similarity metric, as an annotator to determine  whether to retain or remove a candidate summary-document pair as a sample of the distant supervision training (DS-retrain),  as the strategy mentioned in \cite{qin2018robust}. 
% further: threshold.
%or \\
%we adopt computing concept similarity as an annotator to determine whether to retain or remove a candidate summary-document pair as the sample for retraining.  The concept similarity is simply calculated by $tf-idf(C_{train}, C_{test})$ \\%We adopt ROUGE$_\text{F1}$, a text similarity metric, as an annotator to determine  whether to retain or remove a candidate summary-document pair as a sample of the distant supervision training (DS-retrain),  as the strategy mentioned in \cite{qin2018robust}. 
% further: threshold.
%or \\
%we adopt computing concept similarity as an annotator to determine whether to retain or remove a candidate summary-document pair as the sample for retraining.  The concept similarity is simply calculated by $tf-idf(C_{train}, C_{test})$ \\

    %  \begin{tabular}{l|c|c|c}
    %  \hline
    % 	Models &\multicolumn{3}{ c }{Novel $n$-gram (\%)}\\ \cline{2-4}
 
\section{Experiments}

\begin{table*}[t]
	\centering
	\caption{ROUGE F1 evaluation results on the Gigaword and ROUGE recall on  DUC-2004 test set. The results with $\dag$ mark are taken from the corresponding papers. Underlined scores are the best without additional optimization.  Bold scores are the best between the two optimization strategies. $\star$ mark indicates the improvements from the baselines to the concept pointer are statistically significant using a two-tailed t-test ($p<0.01$).
	} 
	\label{tab:test} 
	\begin{tabular}{l|ccc|ccc}
		\hline
		Models & \multicolumn{3}{ c }{Gigaword}&\multicolumn{3}{ c }{DUC-2004}\\\cline{2-4} \cline{5-7}
		&RG-1            & RG-2     & RG-L   & RG-1            & RG-2     & RG-L\\ 
		\hline
	%	ABS$^\dag$   \cite{DBLP:conf/emnlp/RushCW15}      & 29.55           & 11.32     & 26.42   & 26.55           & 7.06     & 22.05    \\
		ABS+$^\dag$  \cite{DBLP:conf/emnlp/RushCW15}    & 29.76           & 11.88     & 26.96  & 28.18           & 8.46     & 23.81       \\
		Luong-NMT$^\dag$  \cite{DBLP:conf/emnlp/LuongPM15} & 33.10   & 14.45 	& 30.71   & 28.55           & 8.79 	& 24.43    \\
		RAS-Elman$^\dag$ \cite{Chopra2016Abstractive} & 33.78 & 15.97 & 31.15 & 28.97 & 8.26 & 24.06\\
	%	RAS-LSTM$^\dag$ \cite{Chopra2016Abstractive} & 32.55 & 14.70	& 30.03 & 27.41 & 7.69	&23.06 \\
		lvt5k-lsent$^\dag$ \cite{DBLP:conf/conll/NallapatiZSGX16}  & 35.30 & 16.64 & 32.62 & 28.61    & 9.42     & 25.24\\
		SEASS$^\dag$   \cite{DBLP:conf/acl/ZhouYWZ17}     & 36.15  & 17.54    &33.63        & 29.21  & 9.56     & 25.51\\
		Seq2seq+att	$^\star$(our impl.)   & 33.11 &	14.67 &	31.06  &28.57 &	9.31 &	24.81 \\
	 	Pointer-generator $^\star$(our impl.) \cite{DBLP:conf/acl/SeeLM17} & 35.98 &	15.99&	33.33&28.28 	&10.04 	&25.69\\
	    Pointer-Cov.-Entail.-Quest.$^\dag$ \cite{DBLP:conf/acl/BansalPG18} & 35.98 & 17.76 &	33.63 & - & - & - \\
	    Seq2seq-Sel.-MTL-ERAM$^\dag$ \cite{DBLP:conf/coling/LiZZZ18} & 35.33 & 17.27 & 33.19 & \underline{29.33} & 10.24& 25.24\\
	    CGU$^\dag$  \cite{DBLP:conf/acl/LinSMS18}    &  36.3 & \underline{18.0} & 33.8 & -&-&-\\
	   % Re$^3$Sum$^\dag$\cite{DBLP:conf/acl/LiWLC18}&	37.04	& \textbf{19.03}&	34.46&-&-&- \\
%	 Concept pointer (top concept) &	36.62 &	16.40 &	33.98 &27.71 &	9.51 &	25.20\\
 %   Concept pointer+RL-ROUGE &	37.93 &	16.70 &	35.32& 28.88 &	10.28 & 26.45\\
  %  Concept pointer+RL-DataDistri. & -&-&-&29.66 &	10.38 &	27.12\\
    \textbf{Concept pointer}  &	\underline{36.62} &	16.40 &	\underline{33.98} &29.17 &	\underline{10.38} &	\underline{26.34}\\ \hline
        \textbf{Concept pointer+RL} &	\textbf{38.02} & {16.97} &	\textbf{35.43} & 29.34 &	9.84 &	26.60\\	
    \textbf{Concept pointer+DS} & {37.01}&\textbf{17.10}&{34.87} & \textbf{30.39} &	\textbf{10.78} & \textbf{27.53}\\
    \hline
	\end{tabular}
	\label{tab:overall}
\end{table*}
\paragraph{Datasets:}
To evaluate the effectiveness of our proposed model, we conducted training and testing on two popular datasets. The first was the English Gigaword Fifth Edition corpus \cite{parker2011english}. % which contains 9.5 million news articles. 
We replicated the pre-processing steps in  \cite{DBLP:conf/emnlp/RushCW15} to obtain the same training/testing data. 
After pre-processing, the corpus contained about 3.8M sentence-summary pairs as training set and 189K pairs as the  development set. 
%We used the same testing set with \cite{DBLP:conf/emnlp/RushCW15} which contained 2000 sentence-summary pairs. 
Once pairs with empty titles were removed, the testing set numbered 1951 pairs. The second dataset, DUC2004, was only used for testing. This dataset consists of 500 document-headline summary pairs, where each document is paired with four reference summaries written by humans.
%This corpus was then produced by pairing the first sentence in the news articles and their headlines as the summaries.  We used the scripts\footnote{\url{https://github.com/facebook/NAMAS/}}released by \cite{DBLP:conf/emnlp/RushCW15} to build the training and development datasets. The script also performed various basic text normalization, including tokenization, lower-casing, replacing all digit characters with $\#$, and marking the words appearing less than 5 times with a UNK tag. 
%Upon the processing, it contained about 3.8M sentence-summary pairs as  training set and 189K pairs as development set. We used the same testing set with \cite{DBLP:conf/emnlp/RushCW15} which contained 2000 sentence-summary pairs. We then removed those pairs with empty titles. The testing set was then slightly changed to 1951 pairs.  The second dataset, DUC-2004 was only used for the testing evaluation. It consisted of 500 document-headline summary pairs, each document was paired with 4 human-written reference summaries which were capped at 75 bytes.
\paragraph{Evaluation Metrics:}
%In the series of experiments, 
We used ROUGE \cite{lin2004rouge} as the evaluation metric, which measures the quality of a summary by computing the overlapping lexical elements between the candidate summary and a reference summary. Following  previous practice, we assessed RG-1 (unigram), RG-2 (bigram) and RG-L (longest common subsequence - LCS). % R-1 and R-2 mainly consider informativeness while R-L is supposed to reveal  readability of the summary.
Noted that the English Gigaword\footnote{The ROUGE evaluation option is, -m -n 2 -s} testing set contains references of different lengths, while the DUC-2004\footnote{The ROUGE evaluation option is, -n 2 -m -b 75 -s} testing set fixes the summary length to 75 bytes. 
%Following previous work, we will report ROUGE F1 for English Gigaward test set, and ROUGE recall for the DUC-2004 test set.

\paragraph{Training Setups:}
%We trained our model on a single GTX TITAN GPU. 
We initialize word embeddings with 128-d vectors  and fine-tune
them during training. Concepts share the same embeddings with the words. The vocabulary size was set to 150k for both the  source and target text. The hidden state size was set to 256. 
The vocabulary size is increased from around 602 to 2216 concepts w.r.t  the different number ($k=1,\cdots, 5$) of concept candidates for each word. 
Note  that the generated concepts with UNKs were subsequently  deleted. 
Our code is available on \url{https://github.com/wprojectsn/codes}, and the vocabularies and candidate concepts are also included. 
% Additional details are provided in Appendix C.

We trained our models on a single GTX TITAN GPU machine.
We used  the Adagrad optimizer with a batch size of 64 to minimize the loss. 
The initial learning rate and the accumulator value were set to 0.15 and 0.1, respectively. We used gradient clipping with a maximum gradient norm of 2. 
At the time of decoding, the summaries were produced through a beam search of size 8.
The hyper-parameter settings were $\lambda = 0.99$, $\gamma=0.1$, $\pi = 2.92$ on  DUC-2004 and $\pi =1.68$ on Gigaword. We trained our concept pointer generator for 450k iterations yielded the best performance,  then took the optimization using RL rewards for RG-L at  95K iterations on DUC-2004 and at 50K iterations on Gigaword. We took the distance-supervised training at 5K iterations on DUC-2004 and at 6.5K iterations on Gigaword.
\paragraph{Baselines:}
The following state-of-the-art baselines were used as comparators. 
%\textbf{ABS} \cite{DBLP:conf/emnlp/RushCW15} used an attentive Convolution Neural Network encoder and Neural Network Language Model   decoder to summarize the sentence.
 \textbf{ABS+} \cite{DBLP:conf/emnlp/RushCW15} is a tuned ABS model with additional features.
 %where the ABS used an attentive CNN encoder and an NNLM decoder.
\textbf{Luong-NMT} \cite{DBLP:conf/emnlp/LuongPM15} is a  two-layer LSTM encoder-decoder.
\textbf{RAS-Elman} \cite{Chopra2016Abstractive} is a convolution encoder and an Elman RNN decoder with attention. 
\textbf{Seq2seq+att} is two-layer BiLSTM encoder and one-layer LSTM decoder equipped with attention. 
\textbf{lvt5k-lsent} \cite{DBLP:conf/conll/NallapatiZSGX16} uses   temporal attention to keep track of the past attentive weights of the decoder and restrains the repetition in later sequences. 
\textbf{SEASS} \cite{DBLP:conf/acl/ZhouYWZ17} includes an additional selective gate to control information flow from the encoder to the decoder.
\textbf{Pointer-generator} \cite{DBLP:conf/acl/SeeLM17}  is an integrated   pointer network and  seq2seq model. 
We implemented this baseline without its coverage mechanism since this is not our focus. 
Baseline models also include two pointer-generator based extensions \cite{DBLP:conf/acl/BansalPG18,DBLP:conf/coling/LiZZZ18}. \textbf{CGU} \cite{DBLP:conf/acl/LinSMS18} sets a convolutional gated unit and self-attention for global encoding.% to select  relevant phrases. 
%\textbf{Re$^3$Sum} retrieved proper summaries as candidate templates, then is extended by  jointly conducting template reranking and template aware rewriting. 

%\paragraph{Testing Questions} We design the experiments particularly to answer the following questions, with the purpose to validate the effectiveness of our proposed model. 

% \begin{enumerate}
%     \item How much percentage of OOV words have been solved. 
%     \item How much percentage of abstract words or concepts has been increased.
%     \item How much percentage of acceptance  has been increased after adopting ConPS.
% \end{enumerate}

\section{Results and Analysis}
%文摘中出现概念词有两种情况：1.文本中多个单词对应一个概念词，这样在beam search时，概念词会因为概率高而被选到候选文摘中；2.文本中没有多个单词对应一个概念词的情况，此时，若是概念词的概率不是很低（经过修正的在概念图中的概率高、概念词对应单词的注意力概率高），那么也有可能被选入到候选文摘中。
The following analysis focuses on investigating whether our model is, first, able to generate abstract and new concepts, and, second, how the overall quality performs against the baselines.
\subsection{Quantitative Analysis}

The results are presented in Table \ref{tab:overall}. We observe that our model outperformed all the strong state-of-the-art models on both datasets in all metrics except for RG-2 on Gigaword. In terms of the  pointer generator performance, the improvements made by our concept pointer are statistically significant 
%using a two-tailed t-test 
($p<0.01$) across all metrics.
% QUESTION: 参照Closed-Book Training to Improve Summarization

\paragraph{OOV and Summary Length:}
\begin{table}[t]
    \centering
    \caption{OOV problem analysis: percentages (\%(NO. UNK/NO. all generated words)) of generating UNK w.r.t the following three models on Gigaword and DUC-2004 datasets.}
    \label{tab:humantest}
    \resizebox{\linewidth}{!}{
    \begin{tabular}{l|c|c}
    \hline
         Method	& Gigaword  & DUC-2004\\
         \hline
         seq2seq+att & 4.02\%(570/14183)	& 2.08\%(85/4079)\\
         Pointer generator &  1.16\%(207/17859)     &	0.31\% (16/5140)    \\
         Concept pointer & 1.12\%(202/17950) &  0.23\%(12/5230)  \\
    \hline
    \end{tabular}
    }
    \label{tab:unk}
\end{table}
 OOV is another major challenge for current abstractive summarization models. Although generating longer summaries or less UNKs is not our focus, our model still showed improvements in this regard (Table \ref{tab:unk}). We counted the number of UNKs and all generated summary words and measured the proportions in both testing sets. 
 The OOV percentages dropped  from 4.02\% to 1.12\% on Gigaword and from 2.08\% to 0.23\% on DUC-2004, which demonstrates that our model is effective at alleviating OOV problems. This result also supports the superior of the concept pointer over the baseline pointer generator. From the statistics, we found that the summaries generated by our concept pointer averaged around 10.46 words, while the pointer generator summaries averaged 10.28 words per summary on DUC-2004. This shows the concept pointer is able to capture more salient content by generating relatively longer summaries. 
 
\paragraph{Abstractiveness:}
 \begin{table}[t]
     \centering
     \caption{Abstractiveness: percentage of novel $n$-grams on Gigaword dataset.}
     \label{tab:abstraction}
     \resizebox{\linewidth}{!}{
     \begin{tabular}{l|c|c|c}
     \hline
    	Models &\multicolumn{3}{ c }{Novel $n$-gram (\%)}\\ \cline{2-4}
         &	$1$-gram &	$2$-gram &	$3$-gram\\ \hline
        Pointer generator &	14.3 &	41.9 &	63.4\\
        Concept pointer  & 17.2 &	45.8 &	68.5\\ \hline
        Reference summary &	25.4 &	65.6 &	78.4\\\hline
     \end{tabular}
     }
     \label{tab:abstract}
 \end{table}
 
 According to \citet{chen2018fast}, abstractiveness scores are  computed as the  percentage of novel $n$-grams in the  generated summaries that are not included in the source documents. As shown in Table \ref{tab:abstract}, compared with human-written summaries which receive the highest novelty in terms of abstractiveness, our concept pointer generator achieves closest  performance with human-written summaries against the baseline. This result demonstrates a further advantage of our model in producing new and abstract concepts. 
Our model is designed to improve semantic relevance and promote  higher abstraction. More generated summary examples can be found in Appendix B.

\subsection{Analysis on Training Strategies}

To evaluate the relative impact of each training strategy with the model, we tested different combinations for comparison with each other and against the baselines. %The results of these models on the testing sets  are shown in Figure \ref{fig:conceptnumber} and Table \ref{tab:overall}.
 
\begin{figure}[t]
    \centering
    \subfigure[Gigaword:  $\widehat{P_i^C}_\text{random}$]
    {\includegraphics[width =3.8cm]{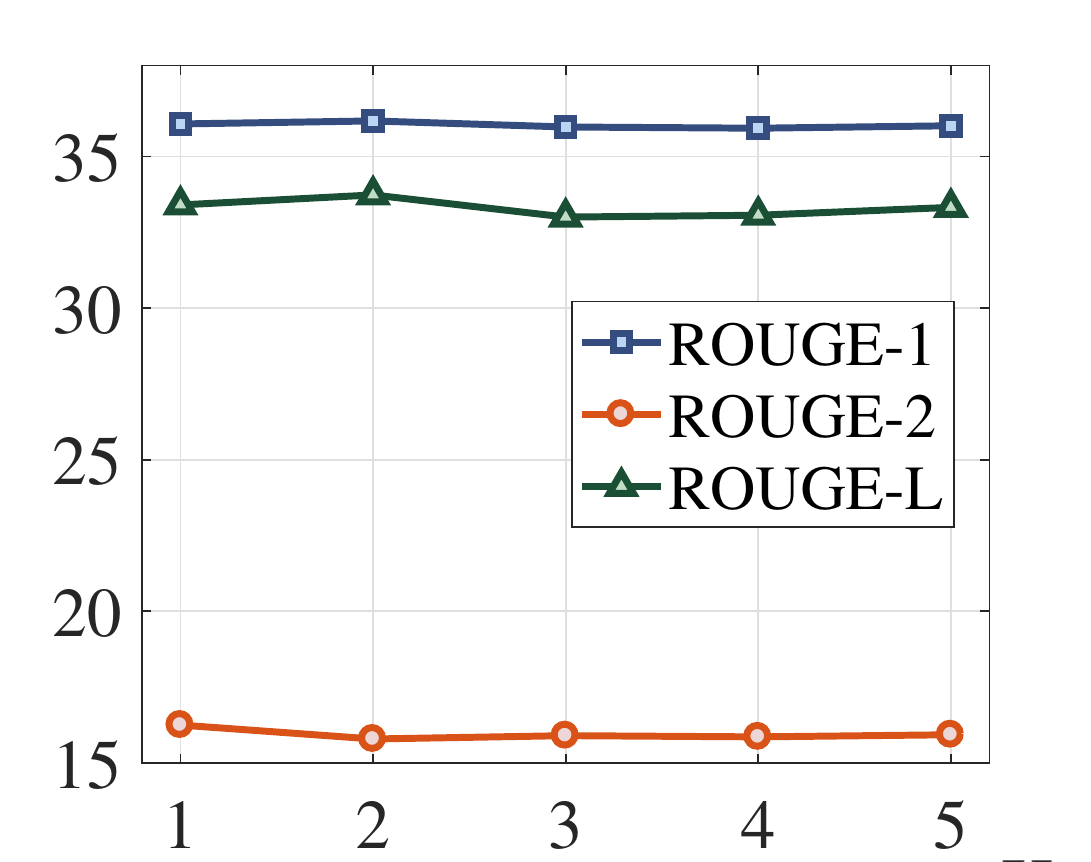}\label{a1}}
    \subfigure[DUC-2004:  $\widehat{P_i^C}_\text{random}$]
    {\includegraphics[width =3.8cm]{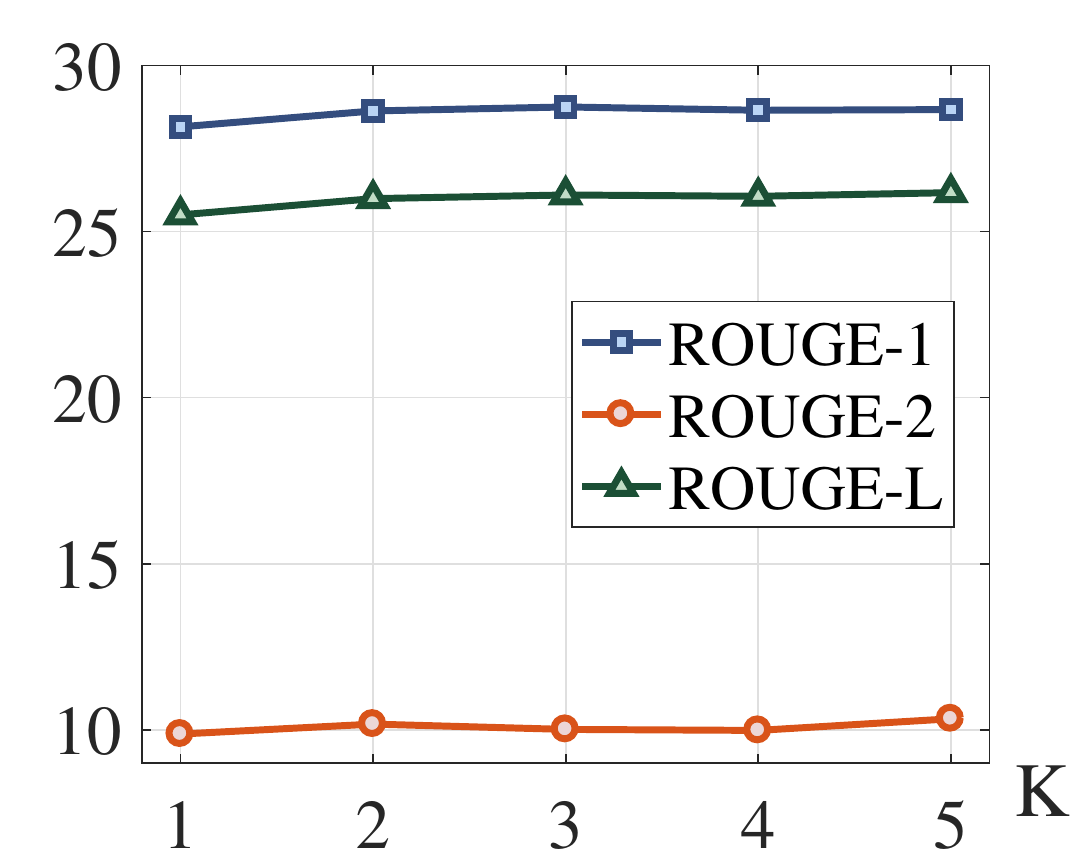}\label{a2}}
    \subfigure[Gigaword:  $\widehat{P_i^C}_{\arg\max}$]
    {\includegraphics[width =3.8cm]{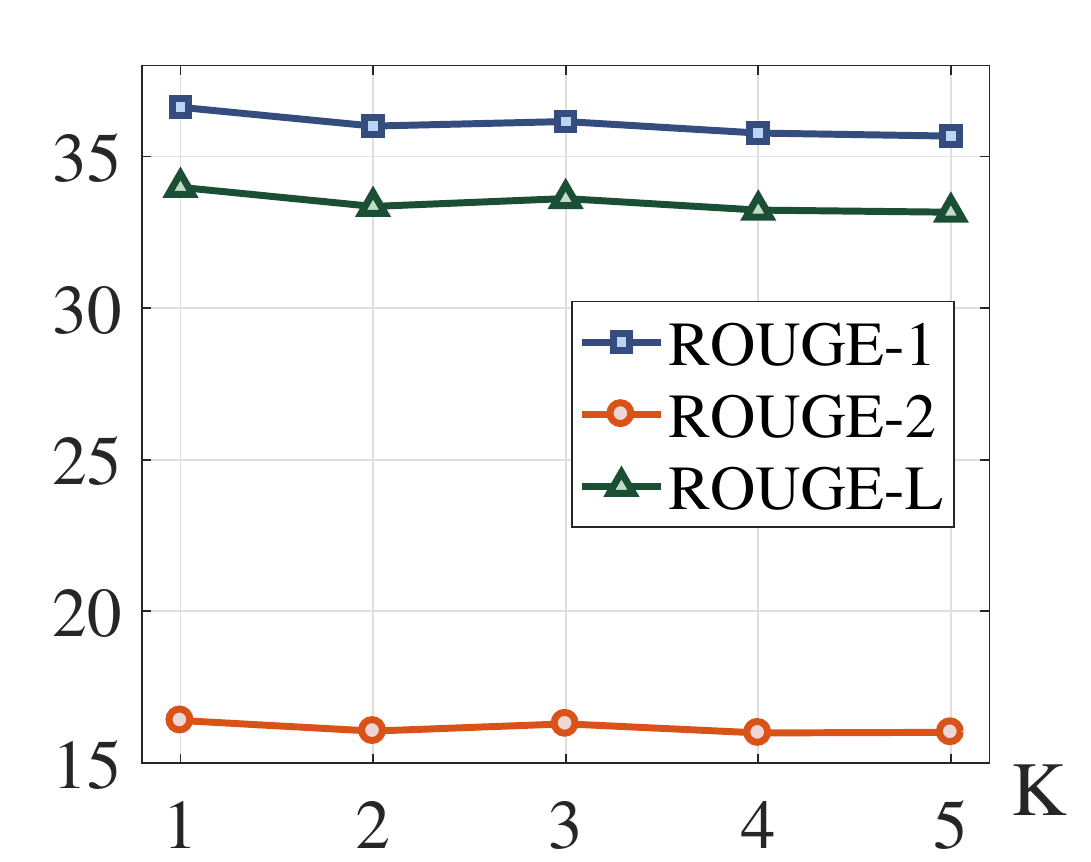}\label{a3}}
    \subfigure[DUC-2004:  $\widehat{P_i^C}_{\arg\max}$]
    {\includegraphics[width =3.8cm]{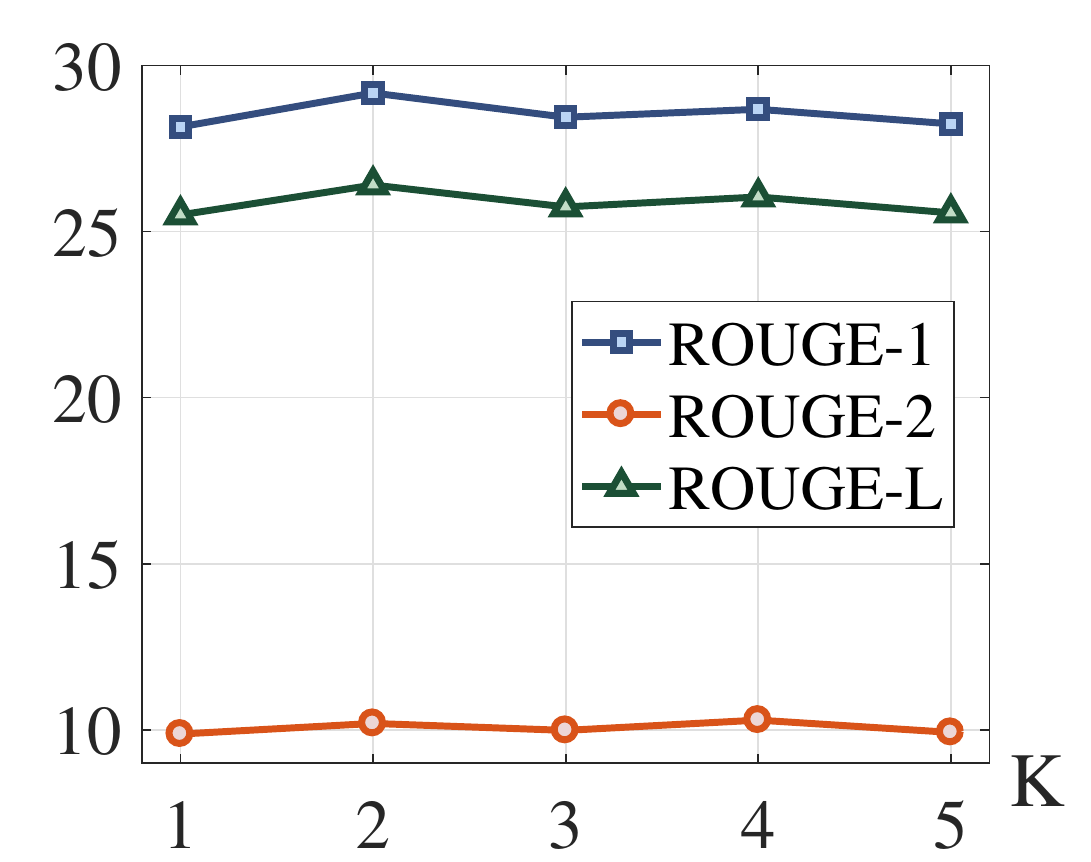}\label{a4}}
    \caption{ROUGE metrics on Gigaword and DUC-2004 w.r.t a different number of concept candidates. Updates were conducted by hard assignment $\widehat{P_i^C}_{\arg\max}$ and random selection $\widehat{P_i^C}_\text{random}$. }
    \label{fig:conceptnumber}
\end{figure}

\paragraph{Context-aware Conceptualization:}
To investigate the impact of training with both the number of concepts $k$ and the concept update strategy mentioned in  Eqs.(\ref{equ:max}) and (\ref{equ:random}), we chose a different number of concept candidates, i.e., $k = 1, 2, 3, 4, 5$, to for the context-aware conceptualization update strategy. Performance was fully evaluated with the three ROUGE metrics as shown in Figure  \ref{fig:conceptnumber}. 
The results only vary slightly according to the number of concepts with the random selection strategy (Eq.(\ref{equ:random})), as shown in Figure \ref{a1} and \ref{a2}. This indicates that a random strategy is not very sensitive to the number of extracted topics.  This is, in part, because the concept pointer may or may not be able to point to the correct concepts from multiple candidates.
While in Figure \ref{a3} and \ref{a4}, the optimum settings are clearly apparent, i.e., $k=1$ on Gigaword and $k=2$ on DUC-2004.  
Overall, the hard assignment strategy (Eq.(\ref{equ:max})) provided the best performance in practical terms, while random selection (Eq.(\ref{equ:random})) performs stably with different settings. 
%This is probably derived by the effect of the proposed  concept pointer that is able to point to the correct concepts from all the candidates. 

%when $ k = 1, 2, 3$ in both datasets; while  $k=4, 5$, the performance decreases. 
%we can find that $k=2$ is the best setting for our model on Gigaword and DUC-2004 datasets.  
%It indicates that our model is not much sensitive to the number of extracted topics. This is probably derived by the effect of the proposed  concept pointer that is able to point to the correct concepts from all the candidates. But, too many concepts also bring noises. 
%that top 2 related concepts extracted from our knowledge base are often  accurate; 
%while more concepts possibly bring noises. 

\paragraph{Training with DS vs. RL:}
% \begin{table}[t]
%     \centering
%     \caption{Ablations: studying impact with different strategies of training concept distribution. Applying RL and distant supervision on the training model.}
%     \label{tab:ablations}
%     \resizebox{\linewidth}{!}{
%     \begin{tabular}{l|c|c|c}
%     \hline
%          Methods	& RG-1 &	RG-2 & RG-L\\
%          \hline
%          Concept pointer ($\widehat{P_i^C}_{\arg\max}$) &  	&   &\\
%          Concept pointer ($\widehat{P_i^C}_\text{random}$) &  	&  & \\
%          +RL+DS-retrain & &	 &\\
%          +DS-retrain+RL& 	& &\\
%     \hline
%     \end{tabular}
%     }
% \end{table}
% As shown in Table \ref{tab:overall}, our model with distant supervision  strategy (concept pointer+DS)  and with reinforcement learning (concept pointer+RL) are all superior to our basic concept pointer generator on both datasets. Furthermore, the relative improvements of concept pointer+DS over concept pointer+RL are ranging from 3.5\% to 9.6\% on DUC-2004, while concept pointer+DS is inferior to concept pointer+RL on Gigaword. The comparing results demonstrate that the DS training plays a noticeable effect when the testing set has semantics differences, but less improvements than RL training if testing set is close to training set. 
%   %Since the model is trained on Gigaword, testing on DUC-2004 indicates the model should be adaptive with different aspects of their contents, called model adaption in this paper.  
% It also verifies the DS strategy is better for model adaption for  abstractive summarization. 
 As shown in Table \ref{tab:overall}, our model with either a distant supervision strategy (concept pointer+DS) or reinforcement learning (concept pointer+RL) were both superior to the basic concept pointer generator on both datasets. Further, the relative improvement of the concept pointer+DS over the concept pointer+RL ranged from 3.5\% to 9.6\% on DUC-2004 but was inferior to concept pointer+RL on Gigaword. In comparing the results, it is clear that DS training has a noticeable effect when the testing set is substantially semantically different from the training set but provides less improvement than RL when the two are close. From this analysis, we conclude that the DS strategy is better for model adaption with abstractive summarization.

\subsection{Human Evaluations}

\begin{table}[t]
    \centering
    \caption{Human evaluation: scoring of three models in terms of abstraction and overall quality by human evaluators (the higher the better). The score range could be 0-20. $\star$ indicates the improvements from the baselines to the concept pointer are statistically significant.}
    \label{tab:humantest}
    \resizebox{\linewidth}{!}{
    \begin{tabular}{l|c|c}
    \hline
         Method	& abstraction &	overall quality\\
         \hline
         seq2seq+att $\star$ & 5.85	& 5.65\\
         Pointer generator $\star$ &8.95&	8.10\\
         Concept pointer &10.00	&9.60\\
    \hline
    \end{tabular}
    }
\end{table}
To explore the correctness of our model using human judgment, we conducted a manual evaluation with 20 post-graduate volunteers. We primarily used the following criteria to assess the generated summaries: {\em abstraction}, i.e., Are the abstract concepts contained in the summary appropriate?; and {\em overall quality}, i.e., Is the summary readable, informative, relevant, etc.? 
%\paragraph{Evaluation Procedure}
To conduct the evaluation, we randomly selected 20 examples from the DUC 2004 testing set and asked the volunteers to subjectively assess the summaries. Each example consisted of an article and three summaries, i.e., a summary by the seq2seq model, the pointer generator model, and our proposed concept pointer model. The volunteers chose the best summaries for each of the articles according to the above criteria (can be multiple choices). Obviously, the summaries were randomly shuffled, and the model used to produce each was unknown to prevent bias. The scores for each model were ranked by how many times the volunteers chose a summary w.r.t each criteria, averaged by the number of participants. The results are presented in Table \ref{tab:humantest}, which show that our model outperformed both the seq2seq model and the pointer generator  \cite{DBLP:conf/acl/SeeLM17} in both criteria. 
 
%Due to the space limitation, we cannot present all the generated cases in the paper. 
%Next, we manually inspect the summaries of our model, According to our observations, we find that it cannot be as abstract as human summaries. Based on the generated summaries, the major behavior of our model is yet viewed as copying segments of source texts and then reorganizing them into a  summary. But it indeed tends to produce more high-level concepts with correct relations that are existed in the source texts comparing to the baselines, additionally produces the long, fluent and informative summary.
 As a last step, we manually inspect the summaries generated by our model, and some examples are presented in Appendix B. We found that the summaries were not as abstract as human-written summary would likely be. The overarching tendency of the model is still to copy segments of the source text and rearrange the phrases into a summary. However, the overall approach does produce more high-level concepts with correct relations compared to the baselines, which demonstrates that our solution is a promising research direction to further pursue. Additionally, the generated summaries are long, fluent, and informative.
 
\section{Conclusion}
% This paper presents a novel concept pointer generator model  to improve the abstractiveness of the model and generate concept-oriented summaries. We also propose a distant supervision for model adaption with a different dataset. Our model achieves  significant improvements over the state-of-the-art baselines on both popular datasets.
This paper presents a novel concept pointer generator model to improve the abstractive summarization model and generate concept-oriented summaries. We also propose a novel distant supervision strategy for model adaption to different datasets. Both empirical and subjective experiments show that our model makes a statistically significant quality improvement over the state-of-the-art baselines on two popular datasets.

\section*{Acknowledgement}
This work was supported by National Natural Science Foundation of China (No.61602036, 61751201), partially supported by the Research Foundation of Beijing Municipal Science \& Technology  Commission (No. Z181100008918002), and partially supported by Ministry of Education China Mobile Research Foundation (No. MCM20170302).
\bibliography{emnlp-ijcnlp-2019}
\bibliographystyle{acl_natbib}

\end{document}